# A MULTI-LEVEL CONVOLUTIONAL LSTM MODEL FOR THE SEGMENTATION OF LEFT VENTRICLE MYOCARDIUM IN INFARCTED PORCINE CINE MR IMAGES

*Dongqing Zhang[1,2#], Ilknur Icke[3], Belma Dogdas[4], Sarayu Parimal[5], Smita Sampath[5], Joseph Forbes[6], Ansuman Bagchi[4], Chih-Liang Chin[5], Antong Chen[1*]*

[1]Applied Mathematics and Modeling, Merck & Co., Inc., West Point, PA, USA.; [2]Department of Electrical Engineering and Computer Science, Vanderbilt University, Nashville, TN, USA; [3]Scientific Informatics, Merck & Co., Inc., Boston, MA, USA; [4]Applied Mathematics and Modeling, Merck & Co., Inc., Rahway, NJ, USA; [5]Translational Biomarkers, Merck Sharp & Dohme, Singapore; [6]High Performance Computing, Merck & Co., Inc., West Point, PA, USA
[#] Contributed as intern at Merck & Co., Inc., West Point, PA, USA.; *antong.chen@merck.com

## ABSTRACT

Automatic segmentation of left ventricle (LV) myocardium in cardiac short-axis cine MR images acquired on subjects with myocardial infarction is a challenging task, mainly because of the various types of image inhomogeneity caused by the infarctions. Among the approaches proposed to automate the LV myocardium segmentation task, methods based upon deep convolutional neural networks (CNN) have demonstrated their exceptional accuracy and robustness in recent years. However, most of the CNN-based approaches treat the frames in a cardiac cycle independently, which fails to capture the valuable dynamics of heart motion. Herein, an approach based on recurrent neural network (RNN), specifically a multi-level convolutional long short-term memory (ConvLSTM) model, is proposed to take the motion of the heart into consideration. Based on a ResNet-56 CNN, LV-related image features in consecutive frames of a cardiac cycle are extracted at both the low- and high-resolution levels, which are processed by the corresponding multi-level ConvLSTM models to generate the myocardium segmentations. A leave-one-out experiment was carried out on a set of 3,600 cardiac cine MR slices collected in-house for 8 porcine subjects with surgically induced myocardial infarction. Compared with a solely CNN-based approach, the proposed approach demonstrated its superior robustness against image inhomogeneity by incorporating information from adjacent frames. It also outperformed a one-level ConvLSTM approach thanks to its capabilities to take advantage of image features at multiple resolution levels.

## 1. INTRODUCTION

In the development of novel therapies for cardiovascular diseases, particularly in the development of a preclinical porcine cardiac disease model, myocardial infarctions are introduced surgically into the specimens to simulate the cardiac pathology, and short axis cine MRI is used as a primary protocol for assessing various structural and functional changes caused by the infarction. The scan is conducted dynamically to cover the complete cardiac cycle at various 2D transverse locations from the base to the apex of the heart, and critical properties e.g. the left ventricle (LV) myocardium thickness, stroke volume and ejection fraction can be obtained based on the segmentation of LV myocardium on the 2D frames. Due to the large number of frames in each scan, manually delineating the myocardium contours is considered a time-consuming and error-prone task. In recent years, various approaches [1, 2] have been proposed for automatic LV myocardium segmentation in human cine MR scans, which could be used on the porcine scans acquired using a similar protocol. Compared with the conventional methods, the latest research [3-5] based primarily on deep convolutional neural networks (CNN) has demonstrated capability in generating highly accurate results with a high level of robustness.

However, majority of the CNN-based approaches tend to treat the frames in the cardiac cycle as independent images, failing to take advantage of the continuous spatial-temporal dynamics of heart motion between consecutive frames. The problem could be critical when segmenting images of infarction, since the diseased specimens can develop myocardial thinning [6], and lesions in the myocardium might cause local intensity decrease [7]. Both problems can lead to observable inhomogeneities in some of the frames of the cardiac cycle, posing more challenges to the segmentation task.

In recent years, recurrent neural networks (RNN), particularly the long short-term memory (LSTM) [8], have been applied successfully for the exploration of dynamic temporal behaviors in areas such as object recognition [9] and natural language processing [10]. For the application of cardiac cine MRI, Poudel *et al.* [5] introduced an RNN at the lowest resolution level of a U-Net CNN to model the spatial continuity between adjacent LV locations. Xue *et al*. [11] proposed a spatial-temporal circle LSTM model to estimate the LV myocardium thicknesses at the six regions of the mid-slice in the short axis scan. Both works succeeded by introducing RNNs to process abstracted features extracted by the corresponding CNN models.

Inspired by these RNN-based approaches, especially the work in [11], we propose a multi-level convolutional LSTM (ConvLSTM) approach for the automatic segmentation of LV myocardium. To develop the ConvLSTM model, a ResNet-56 CNN model [12] is trained first, and the LV-related image features at the low- and high-resolution levels are extracted separately, each for training one LSTM model. Using a leave-one-out approach, we compare the proposed model with a one-level ConvLSTM and a CNN model by evaluating them in a dataset of which the LV myocardium is manually delineated. We use Dice similarity coefficient (DSC), Hausdorff distance (HD), and average perpendicular distance (APD) compared with the manual delineation as measures of accuracy. The performance of the proposed multi-level model is found to be the best among the three.

## 2. METHOD AND MATERIALS

### 2.1. Image preprocessing

Both training and testing images are preprocessed following the steps described in [12], where each slice is resampled to $1\times1$ mm$^2$ pixel size and cropped to a $184\times184$ matrix from the center. Further LV localization is performed by motion analysis and Hough transform for detecting circles, such that the bounding box enclosing the LV area can be defined with a set margin. For the purpose of training and testing the LSTM, the cropped frames are grouped and ordered by cardiac cycles such that each group consists of the $N$ ($N = 25$ in our application) frames in their original temporal order.

### 2.2. Neural network architecture

The overall architecture of the networks is shown in Fig. 1 (a), which is mainly a combination of two major network blocks, i.e. the CNN block and the RNN block, that are trained separately with the same training data.

*2.2.1. CNN block*
The CNN block consists of a fully convolutional 56-layer residual learning network [13] constructed with building units shown in Fig. 1(b). Inside a building unit, a shortcut would allow information to pass directly to the output, which allows the convolutional layers to learn the residual to the unit's input and alleviates the gradient diminishing problem in training deep CNNs.

After feeding the training images into the CNN blocks, the generated feature maps are downsampled by a factor of 2 (controlled by convolution stride size) whenever the data are processed by 9 building units. The final outputs of each downsampled resolution level (by 2 and 4 herein) are used as the primary feature maps for training the RNN. Note that since the CNN block is mainly used as a feature extractor, any CNN architecture that is able to extract image features effectively can be used to replace the ResNet architecture here. To free up GPU resource, the training of the CNN block is conducted ahead of training the RNN block.

*2.2.2. RNN block*
The RNN block consists of two levels of ConvLSTM blocks, each of which is dedicated to processing the image feature maps at a corresponding resolution level. In each block, as it is shown in Fig. 1(c), a series of LSTM units with length $N_{LSTM}$ which is the number of frames in one cardiac cycle are serially connected.

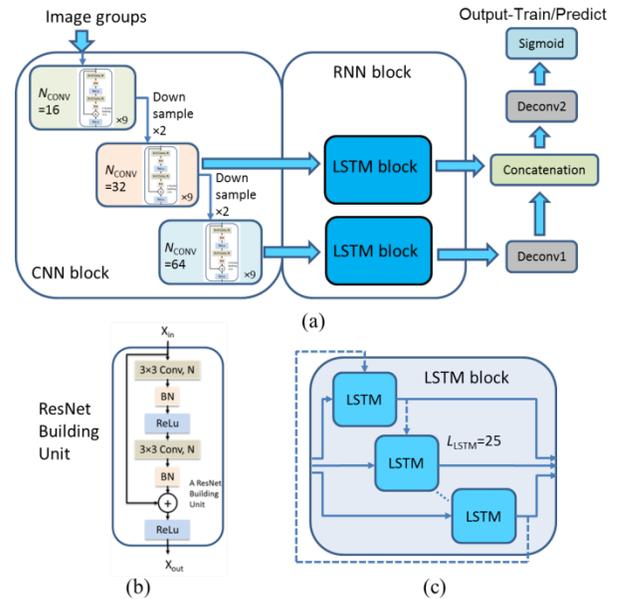

Fig. 1. The architecture of the proposed multi-level LSTM model, which shows (a) the overall structure, (b) the ResNet building unit used in the CNN block, and (c) the infrastructure of an LSTM block used in the RNN block.

Each of the LSTM unit has an infrastructure as it is shown in Fig. 2. The internal gates and the output are updated following equations (1) to (5):

$$i_t = \sigma(W_{xi} * X_t + W_{hi} * H_{t-1} + W_{ci} \circ C_{t-1} + b_i) \quad (1)$$

$$f_t = \sigma(W_{xf} * X_t + W_{hf} * H_{t-1} + W_{cf} \circ C_{t-1} + b_f) \quad (2)$$

$$C_t = f_t \circ C_{t-1} + i_t \circ \phi(W_{xc} * X_t + W_{hc} * H_{t-1} + b_c) \quad (3)$$

$$o_t = \sigma(W_{xo} * X_t + W_{ho} * H_{t-1} + W_{co} \circ C_t + b_o) \quad (4)$$

$$H_t = o_t \circ \phi(C_t) \quad (5)$$

where $i_t$ is the input gate, $f_t$ is the forget gate controlling the memory, $o_t$ is the output gate, and $C_t$ is the cell state. Symbol * represents the convolution operation, and ∘ represents the matrix element-wise product. Function $\sigma$ calculates element-wise sigmoid activation, and $\phi$ calculates element-wise hyperbolic tangent activation. Matrix $X_t$ stands for the input fed from external sources, $H_t$ is the hidden state carried over to the next LSTM and used for generating the output. Trainable parameters are $W$'s as weight matrices and $b$'s as bias terms. Compared with conventional LSTM [8], ConvLSTM replaces the matrix vector multiplication with the convolution operation, which significantly reduces the number of parameters to learn while capturing the spatial relation between voxels more effectively [14]. Similar to the circle LSTM in [11], the hidden state of the last LSTM unit in the series is connected to the first unit to reflect the periodic motion of LV.

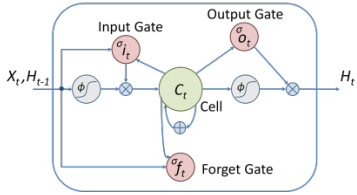

Fig. 2. Structure of a ConvLSTM unit.

*2.2.3. Generation of segmentation via deconvolution*
The features as the output tensor of the LSTM block at the lower resolution level are deconvoluted and concatenated with the output tensor of the LSTM block at the higher resolution level. The concatenated tensor is deconvoluted for another time and then goes through a sigmoid activation to obtain the segmentation result as a probability map, where each pixel is assigned a value between 0 and 1 indicating the probability for the pixel to belong to the myocardium. By combining features at multiple resolution levels, features at both refined and coarse levels are able to contribute to the generation of the segmentation results.

## 3. EXPERIMENTS AND RESULTS

A leave-one-out experiment was carried out on a set of porcine cardiac cine MR images acquired in house. The images were collected on 8 porcine specimens, each of which was scanned at three time points with the first being the baseline and the other two scanned after a surgical induction of cardiac infarction. Each scan included 13 consecutive short axis locations on the entire heart at 25 frames per cardiac cycle, and only the locations covering the LV were selected, leading to a total of 3,600 2D slices included in the experiment. All intervention and imaging experiments were reviewed and approved by the IACUC of Merck & Co., Inc. (West Point, PA, USA) and the National University of Singapore.

In each run of the leave-one-out experiment, images of one specimen (450 slices) were set for testing, and the remaining images were grouped by cardiac cycles that were randomly divided into training and validation sets at an 80/20 ratio. An Adam optimizer [15] was used in the training to minimize the cross-entropy loss. Both the training and testing were implemented using the Tensorflow deep learning framework (https://www.tensorflow.org/) with Python 3 and executed on NVIDIA Tesla P100 GPUs.

To reduce the effect of model variabilities, 5 CNN models were trained using different random initializations, and 5 RNN were trained by initializing from the 5 CNN models respectively. In testing, the 5 models were used to generate 5 segmentations that were averaged and thresholded at intensity of 0.5 out of 1 (1 is the maximum). The final segmentations were compared with the manual segmentations using Dice similarity coefficient (DSC) [16], Hausdorff distance (HD) [17], and average perpendicular distance (APD) [18] from the automatic to the manual segmentations as measures of accuracy, which are reported in Table 1. In addition, results were also obtained using a ResNet-56 CNN model, as well as a ConvLSTM model trained with features at the lowest resolution level only, and compared with the manual segmentations. As it is shown in Table 1, the proposed multi-level LSTM model was able to produce higher DSC and lower HD and APD on average for the 8 experiments compared with the other two methods.

Fig. 3(a) shows an example in which the lower region (yellow arrow) suffered from myocardial thinning due to infarction. The segmentation by the CNN approach collapsed at this region. Although both ConvLSTM-based approaches were able to achieve reasonable results, the multi-level ConvLSTM outperformed slightly. Fig. 3(b) shows an example where the right side (yellow arrow) has relatively lower intensity due to a lesion. The CNN-based segmentation was particularly irregular at the region, which was also very different from the contours on the previous and following frames. The multi-level ConvLSTM effectively corrected the irregularity with the help of adjacent frames, while the one-level ConvLSTM approach had an incomplete contour as a result of error caused by lack of resolution in both the current and previous frames.

## 4. DISCUSSION AND CONCLUSION

A multi-level ConvLSTM model is introduced for the automatic segmentation of LV myocardium in the short axis cine MR images of porcine specimens with myocardial infarction. The proposed approach outperformed a CNN model and a one-level ConvLSTM model in a leave-one-out experiment involving 8 specimens with surgically induced infarction, mostly due to its capabilities in overcoming image inhomogeneities by capturing the spatial-temporal characteristics of image series at multiple resolution levels.

Since the dataset includes only 144 cardiac cycles from 8 subjects, which makes it difficult to evaluate the significance of improvements brought by the proposed model compared with other approaches, we are collecting more images and LV myocardium delineations from additional specimens to extend the evaluation. We hope that by expanding the training dataset the generalizability of the model could be further improved. Transfer learning based on well-labelled human public datasets is also planned.

Table 1. The averaged mean and standard deviation (in brackets) of slice-wise DSC, HD, and APD of automatic VS. manual segmentations for the 8 experiments under the leave-one-out framework.

| Model | CNN | One-level ConvLSTM | Multi-level ConvLSTM |
|---|---|---|---|
| DSC | 0.840 (0.081) | 0.839 (0.083) | 0.842 (0.080) |
| HD (mm) | 6.784 (4.544) | 6.355 (4.285) | 6.125 (4.042) |
| APD (mm) | 1.110 (0.564) | 1.086 (0.556) | 1.037 (0.457) |

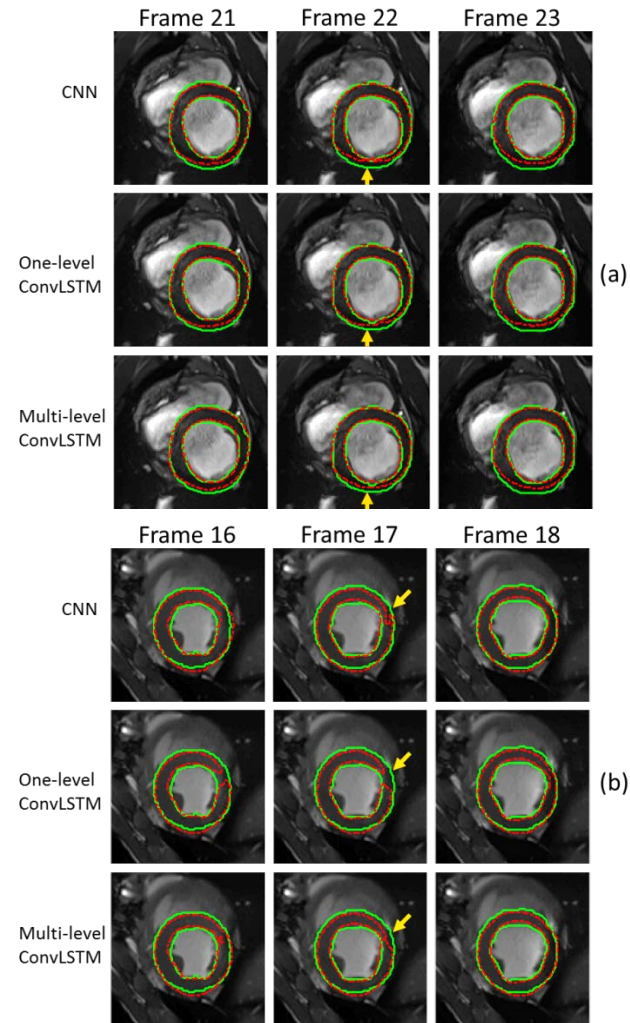

Fig. 3. Segmentation contours overlaid on top of consecutive image frames. Manual contours are in green solid lines and automatic contours are in red dash lines. The middle columns show the frames to be focused on, and the left and right columns show the preceding and succeeding frames, respectively.